\pdfoutput=1

\documentclass[11pt]{article}

\usepackage{naacl2021}

\usepackage{times}
\usepackage{latexsym}

\usepackage[T1]{fontenc}

\usepackage[utf8]{inputenc}

\usepackage{microtype}

\usepackage{enumitem}
\setlist{nosep, leftmargin=5mm} 

\usepackage{letltxmacro}
\LetLtxMacro\oldttfamily\ttfamily
\DeclareRobustCommand{\ttfamily}{\oldttfamily\csname ttsize\endcsname}
\newcommand{\setttsize}[1]{\def\ttsize{#1}}%

\title{Conversational Answer Generation and Factuality for Reading Comprehension Question-Answering}

\author{Stan Peshterliev \\ \texttt{stanvp@fb.com } \\\And Barlas Oguz \\ \texttt{barlaso@fb.com} \\\And Debojeet Chatterjee \\ \texttt{debo@fb.com} \\\And Hakan Inan \\ \texttt{inan@fb.com} \\\And Vikas Bhardwaj \\ \texttt{vikasb@fb.com}
    \\}

\begin{document}
\maketitle
\begin{abstract}
Question answering (QA) is an important use case on voice assistants. A popular approach to QA is extractive reading comprehension (RC) which finds an answer span in a text passage.  However, extractive answers are often unnatural in a conversational context which results in suboptimal user experience. In this work, we investigate conversational answer generation for QA. We propose AnswerBART, an end-to-end generative RC model which combines answer generation from multiple passages with passage ranking and answerability. Moreover, a hurdle in applying generative RC are hallucinations where the answer is factually inconsistent with the passage text. We leverage recent work from summarization to evaluate factuality. Experiments show that AnswerBART significantly improves over previous best published results on MS MARCO 2.1 NLGEN by 2.5 ROUGE-L and NarrativeQA by 9.4 ROUGE-L. 
\end{abstract}

\setttsize{\small}%

\section{Introduction}

Question answering (QA) based on reading comprehension (RC) aims to answer a question from text paragraph. RC is studied in two settings: extractive and generative. In extractive RC, the answer is marked as a span in the text paragraph, whereas in generative RC, the answer is free-form text which requires natural language generation (NLG). Recently, researchers made significant progress in extractive RC using pretrained language models such as BERT~\citep{devlin2018bert} and RoBERTa~\citep{liu2019roberta}. Furthermore, pretrained seq2seq generative models such as BART~\citep{lewis2019bart} and T5~\citep{raffel2019exploring} have been successfully applied to extractive RC~\citep{lewis2020retrieval,izacard2020leveraging}. However, their capabilities for generative RC answers is less studied.


In this paper, we investigate generative RC with pretrained seq2seq models for conversational answer generation. Answers are conversational if they are natural and suitable for voice assistants such as Alexa, Siri and Google Assistant. Conversational answers improve the QA user experience on voice assistants. For example, for the question ``Who is Drake's father?'' an extractive RC model returns  ``Dennis Graham'' from the sentence ``The cover art of Drake's project features a photo of \underline{Dennis Graham}, his father''. In contrast, a conversational answer is ``Drake's father is Dennis Graham.''

A hurdle in working with generative models is that they can produce text that is factually inconsistent with the source information ~\citep{marcus2020next}. Recent work on summarization proposes natural language inference (NLI) entailment as a framework for evaluating factuality of generated text. Factuality is important for QA because we want the answer to be grounded in the input passages. From a user perspective, giving the correct answer may not be sufficient without supporting evidence from the input passage. Even if the RC model hallucinated the correct answer, the user may not accept it. Thus, we consider factuality with respect to the input passages an important metric for generated conversational answers.

\begin{table}
\centering
\small
\setlength{\tabcolsep}{0.5em} 
{\renewcommand{\arraystretch}{1.5}
\begin{tabular}{p{0.3\linewidth} | p{0.60\linewidth}}
\hline
Question & How many terminals are at JFK? \\
Gold Passage & With six terminals, the airlines that serve JFK airport are spread across. \\
Extractive Answer & six terminals \\
Conversational Answer & There are six terminals at the JFK airport. \\
\hline
\end{tabular}
}
\caption{\label{examples}
Different types of answers in MS MARCO
}
\vspace{-6mm} 
\end{table}

To generate conversational answers, we propose AnswerBART, an end-to-end seq2seq model based on BART. Unlike prior work, AnswerBART fully leverages the seq2seq approach to learn to rank, abstain from answering, and generate answers from multiple passages and the question itself. Experiments demonstrate state-of-the-art results on the MS MARCO 2.1 NLGEN task and NarrativeQA. 

Our main contributions are as follows:
\begin{itemize}
\item An end-to-end model for conversational answer generation with reranking and abstaining.
\item Factuality evaluation framework based on NLI entailment which measures correctness and detects hallucinations.
\end{itemize}





\section{AnswerBART}

AnswerBART is an end-to-end conversational answer generation model based on BART. BART is a seq2seq transformer~\citep{vaswani2017attention} pretrained on a large text corpus as a denoising autoencoder. One of the noising functions in BART is shuffling text which has to be reconstructed. This resembles the generative QA task which roughly rearranges the question and passages into an answer. 

\textbf{Problem setting}. We are given a question $q_i$ and a set of passages $\{p^i_j\}$ that are returned from a retrieval system~\citep{karpukhin2020dense}. The goal is to generate a conversational answer $a_i$.

\textbf{Mixed Style Training}. Annotations for QA datasets are often based on extractive answers and a small portion of the dataset is annotated with generative answers. Mixed style training~\citep{nishida2019multi} (MST) uses both extractive and generative answers to train a single RC model and controls for the answer type using a style token. We use MST to leverage the full training data in MS MARCO.

\textbf{Source.} The source sequence consists of a style token, a question, and a set of passages with the separator token \texttt{</s>} in between. The style token can be either \texttt{s:extract} or \texttt{s:conv} to indicate an extractive or conversational answer. The question is prefixed with \texttt{q:} and each passage is prefixed with an index \texttt{p$_{j}$:}. For example, ``\texttt{s:conv} \texttt{</s>} \texttt{q:} $q_i$ \texttt{</s>} \texttt{p0:} $p_i^0$ \texttt{</s>} \texttt{p1:} $p_i^1$ \texttt{</s>} \texttt{p2:} $p_i^2$ \texttt{</s>}''. 

\textbf{Target.} The target sequence consists of a passage ranking segment and a generated answer segment. For example, “\texttt{p1:} \texttt{p2:} \texttt{p0:} $a_i$”. We strip the ranking and use only the answer in our evaluations.

\textbf{Passage Ranking}.  The passage ranking based on $n$-gram algorithms like BM25 can be suboptimal for QA. The reason is that BM25 does not consider the rich semantics between the query and a passage. To improve ranking, we use a transformer passage reranker where the self-attention over the query and a passage captures semantic interactions~\citep{nogueira2019passage}. However, to avoid having an additional model at runtime, we prepend the passage ranking to the target sequence which makes AnswerBART learn the ranking. This allows AnswerBART to attend more to passages that are more likely to contain an answer.

\textbf{Answerability}. If the passages do not contain an answer, the model generates ``No Answer Present''. Earlier work on extractive QA returns a 0 length span or uses an additional head for answerability.

\subsection{Model Extensions}

We evaluated the addition of two extensions to enhance AnswerBART. Both did not improve the base AnswerBART architecture which we show in the ablation study section.

\textbf{Copy pointer}. We add a copy pointer head to allow AnswerBART to copy tokens directly from the query and the passages. We follow \citet{einolghozati2020sound} which first initializes the copy head with the average of the last layer’s pretrained decoder attention head, and adds a loss that forces the decoder to use the copying mechanism.

\textbf{Combination with extractive RC}. We use an extractive model to get an answer span from a passage. Then, we either  (1) use the extracted \textit{answer-only}, e.g., ``\texttt{s:conv} \texttt{</s>} \texttt{q:} albany mn population \texttt{</s>} \texttt{p0:} 2,662'', or (2) mark the \textit{answer-span} in the passage with \texttt{<a></a>}, e.g., ``\texttt{s:conv} \texttt{</s>} \texttt{q:} albany mn population \texttt{</s>} \texttt{p0:} Albany, Minnesota has a community population of \texttt{<a>}2,662\texttt{</a>} people.''. \citet{tan2017s} use a similar approach.

\subsection{Related Models}

AnswerBART closely follows the Masque model~\citep{nishida2019multi} . Masque uses a pretrained ELMo encoder and a randomly initialized transformers decoder with copy pointer and separate prediction heads for ranking and answers detection. AnswerBART uses a transformer for both the encoder and decoder which are initialized with the pretrained BART weights.

%
RAG~\citep{lewis2020retrieval} is a similar QA model based on BART. The difference with our work is that AnswerBART focuses on conversational answer generation with passages from the given task, and RAG focuses on open domain QA with passage retrieval from Wikipedia.


\section{Factuality and Correctness}

Generative models can hallucinate text that is factually inconsistent with the source information~\citep{marcus2020next} which is undesirable for production deployment. To measure factuality in generative RC, we adopt the NLI framework proposed for summarization by \citet{marcus2020next}. The NLI task is given a premise and a hypothesis to predict if the hypothesis entails, contradicts, or is neutral. We consider two methods of applying NLI to QA factuality evaluation.

\textbf{NLI on Passage.}  The passage is the NLI premise and the generated answer is the hypothesis. A factual answer without hallucinations entails from the passage. For multiple passages, we require the answer to entail from at least one passage. Note that NLI models built on datasets such as MultiNLI~\citep{N18-1101} are suboptimal for QA. In MultiNLI, both the premise and the hypothesis are sentences whereas the QA passages can be multiple sentences with contrasting opinions.

\textbf{NLI on Answer.} First, the gold answer is the premise and the generated answer as the hypothesis. Then, we do inference in the opposite direction, the generated answer is the premise and the gold answer is the hypothesis. A correct answer should entail in both directions. The bi-directional entailment is necessary, otherwise the generated answer may not fully answer the question or contain extra information.

\section{Experiments}

\textbf{Datasets.} MS MARCO 2.1~\citep{bajaj2018ms} is a QA dataset built from Bing search logs which consists of queries, ten candidate passages per query, and answers. We focus on the NLGEN subset that is annotated with conversational answers. Note that we are unable to evaluate on the test set because the MS MARCO team stopped receiving submissions for the QA tasks. Thus, we take a random 14K examples subset of the training set as a validation set and use the development set for final evaluation.

In addition, following \citet{nishida2019multi}, we evaluate on NarrativeQA~\cite{kovcisky2018narrativeqa} which is not conversational but it is specifically built for generative answers with reasoning over long passages. 

\begin{table}[ht]
\small
\setlength{\tabcolsep}{0.5em} 
{\renewcommand{\arraystretch}{1.3}
\begin{tabular}{llll}
Dataset & Train & Dev & Test \\ \hline
MS MARCO ALL & 808,731 & 101,093 & 101,092\** \\ 
MS MARCO ANS & 503,370 & 55,636 & - \\ 
MS MARCO NLGEN & 153,725 & 12,467 & - \\ 
NarrativeQA & 32,747 & 3,461 & 10,557 
\end{tabular}
}
\caption{\label{dataset_sizes}
 MS MARCO 2.1 and NarrativeQA dataset statistics. ANS are the answerable questions and NLGEN are the conversational answers. \**no public answers on the test set.
}
\vspace{-4mm} 
\end{table}

\textbf{Metrics.} BLEU (B-1, B-2), ROUGE (R-L), and Meteor (M) metrics are standard for evaluating MS MARCO 2.1 and NarrativeQA. We use the percentage of NLI entailment on Passage (N-P) and on Answer (N-A) for measuring factuality.

\textbf{Setup.} We evaluate the two official BART versions: base with 6 encoder and decoder layers and 786 embedding size, and large with 12 encoder and decoder layers and 1024 embedding size. We implement our model in PyText~\cite{aly2018pytext}. We finetune using the stochastic weight averaging optimizer \cite{izmailov2018averaging} and cross entropy loss with batch size of 16 for 5 epochs. We run all experiments 5 times and average the results.

For passage reranking, we use an ELECTRA-large~\citep{clark2020electra} finetuned on MS MARCO 2.1 using pairwise ranking loss~\citep{nogueira2019passage}. For the combination experiments, the extractive RC model is ELECTRA-large finetuned on SQUAD 2.0~\citep{rajpurkar2018know}.

\subsection{Results}

\textbf{MS MARCO.} Table~\ref{ms_main_results} shows the MS MARCO results together with Masque, the previous best published model. All models were trained using MST. AnswerBART-large improves upon Masque by 2.54 ROUGE-L and 4.83 BLEU-1 points. The margin between AnswerBART-base and Masque is small which suggests that bigger model size is important for improving performance. It is notable that AnswerBART-base captures information about answer generation and passage ranking as well as Masque, whereas the latter has specialized prediction heads and loss functions to capture the diverse information.

\begin{table}[ht]
\small
\centering
\setlength{\tabcolsep}{0.5em} 
{\renewcommand{\arraystretch}{1.3}
\begin{tabular}{lcccc}
Model & B-1 & R-L & N-P & N-A  \\ \hline
AnswerBART-large MST & \textbf{70.39} & \textbf{72.31} & \textbf{97.20} & \textbf{48.43} \\
AnswerBART-base MST & 66.52 & 70.29 & 96.65 & 46.12 \\ \hline
Masque (NLG) & 65.56 & 69.77 & - & -   
\end{tabular}
}
\caption{\label{ms_main_results}
MS MARCO 2.1 NLGEN results on the development set since test set submissions are no longer accepted.
}
\vspace{-4mm} 
\end{table}

\textbf{Factuality and Correctness.} The generated answers are factual with respect to the passage (N-P) with only 3-4\% of answers that do not entail. However, we obtain only 48-46\% of answer entailment (N-A) which means generated answers are often incomplete or contain redundant information with respect the gold answers. Thus, AnswerBART represents the source information correctly but tends to focus on the wrong information to answer the question.

We did manual analysis of hallucination for 100 samples. We found that AnswerBART frequently tries to expand acronyms like state and organization names. Also, AnswerBART hallucinates states for cites, e.g., a passage about a city in Arizona had an answer with the city being in Florida.

\textbf{Answerability.} AnswerBART-base MST detects answerable questions well with F1 score of 78.97 compared to Masque F1 score of 78.93.

\textbf{NarrativeQA.} Table~\ref{narr_main_results} shows the NarrativeQA results. Again, AnswerBART-large achieves significant improvement over Masque by 9.43 ROUGE-L and 13.04 BLEU-1 points. For NarrativeQA, we try to incorporate MST by adding MS MARCO conversational data. However, the results degrade because of data imbalance since NarrativeQA data is five times smaller than MS MARCO data. If we subsample the MS MARCO data to balance the two datasets the results improve but they are still worse than using just NarrativeQA data.

\begin{table}[ht]
\small
\setlength{\tabcolsep}{0.5em} 
{\renewcommand{\arraystretch}{1.3}
\begin{tabular}{p{0.42\linewidth}cccc}
Model & B-1 & B-4 & M & R-L \\ \hline
AnswerBART-large & \textbf{67.15} & \textbf{39.99} & \textbf{33.33} & \textbf{69.30} \\
\hspace{1mm}w/ MST & 62.34 & 36.50 & 30.66 & 66.76 \\
\hspace{1mm}w/ balanced MST  & 66.46 & 39.31 & 33.04 & 68.88 \\
(development set) & 64.96 & 36.54 & 32.31 & 69.19 \\ \hline
Masque & 54.11 & 30.43 & 26.13 & 59.87 
\end{tabular}
}
\caption{\label{narr_main_results}
NarrativeQA results. We use the summaries task which has a single passage.
}
\vspace{-4mm}
\end{table}

\subsection{Ablation and Factuality}

To better study our model, we perform ablations on the MS MARCO 2.1 dataset. We use the AnswerBART-base model as it is faster to train.

\textbf{Ranking.} We experiment with passage ranking, see Table~\ref{ranked}. The \textit{Ranked} $N$ models use the ELECTRA passage ranker for both training and runtime where $N$ are the top passages passed to the model. The \textit{End-to-end} model uses the ELECTRA passage ranker at training time to add ranking segments in the target but it does not need it as runtime. Also, the \textit{No ranking} model uses the passage order from the dataset. 

The ranking ablations show that using multiple passages improves performance. If we use the top ranked passage the accuracy drops because the passage ranker sometimes misses the correct passage. Increasing the number of passages from the 1 to 5 improves performance. However, the improvements saturate beyond 5 because the ranker top-5 accuracy is 99\%. Using no ranking hurts accuracy. Finally, the end-to-end model is able to learn the ranking order and has almost identical results to Ranked 10. Thus, we don't need the ranker at runtime to achieve good performance.

\begin{table}[ht]
\centering
\small
\setlength{\tabcolsep}{0.5em} 
{\renewcommand{\arraystretch}{1.3}
\begin{tabular}{lcccc}
Model & B-1 & R-L & N-P & N-A \\ \hline
Ranked 1 & 66.54 & 68.84 & 90.28 & 42.74  \\
Ranked 5 & 68.53 & 70.27 & 96.22 & 46.09 \\
Ranked 10 & \textbf{66.54} & \textbf{70.30} & \textbf{96.70} & \underline{46.11} \\
No ranking & 66.12 & 70.01 & 96.46 & 45.32 \\
End-to-end & \underline{66.52} & \underline{70.29} & \underline{96.65} & \textbf{46.12}
\end{tabular}
}
\caption{\label{ranked}
Ranking ablations on MS MARCO 2.1 with AnswerBART-base and different number of passages.
}
\vspace{-4mm}
\end{table}

\textbf{Modeling Configurations.} We experiment with different modeling configurations using golden passages without ranking, see Table \ref{configuration_results}. The results are much better than using multiple passages. So, there is a significant potential benefit in improving the ranking accuracy. 

MST provides consistent improvements to BLUE and ROUGE scores as well as factuality. The factuality improves because with the addition of extractive data the model learns to copy tokens from the input instead of generating new tokens. The addition of copy pointer and copy loss does not significantly change the results. The reason is that AnswerBART is already good at copying without a pointer head and uses BPE vocabulary which avoids unknown tokens. Using a combination with an extractive reader does not yield improvements for answer-span, and for answer-only results in large degradation because the model does not have enough context to generate an answer.

\begin{table}[ht]
\small
\setlength{\tabcolsep}{0.5em} 
{\renewcommand{\arraystretch}{1.3}
\begin{tabular}{p{0.4\linewidth}cccc}
Model & B-1 & R-L &  N-P  & N-A \\ \hline
End-to-end  & 81.32 & 79.92  & 93.87 & 62.10 \\
\hspace{1mm}w/ MST & \textbf{82.39} & \textbf{81.01} & \textbf{94.56} & \textbf{64.73} \\
\hspace{1mm}w/ MST copy & 82.36 & 80.86 & 94.26 & 64.38 \\
\hspace{1mm}w/ MST copy loss & 81.85 & 80.67 & 94.46 & 63.95 \\
Answer-span MST & 81.84 & 80.71 & 94.39 & 64.67 \\
Answer-only MST & 66.91 & 71.09 & 72.79 & 46.55
\end{tabular}
}
\caption{\label{configuration_results}
Modeling ablations on MS MARCO 2.1 with AnswerBART-base and golden passages without ranking.
}
\vspace{-4mm}
\end{table}

\textbf{NLI Metrics Analysis.} We randomly sampled 100 development set predictions from the AnswerBART model on gold passages. Then, we asked human annotators to judge them in two rounds. First round, judge if the generated answer factually represents the passage. Second round, judge if the generated answer is correct and complete according to the gold answer. We used the ratings from the first round to validate the NLI on Passage metric (N-P) and the ratings from the second round to validate the NLI on Answer metric (N-A). 

Table~\ref{nli_metric_results} shows the results from the human annotation along with the automated metrics. Both N-P and N-A metrics have high level of agreement with human annotation where N-P agrees with N-P Human for 90\% of the examples and N-A agrees with N-P for 94\% of examples. For the examples where there was disagreement between NLI and human annotation, the error was in the NLI model which suggest that fine-tuning on QA data can further improve the NLI model for factuality and correctness.

\begin{table}[!ht]
\small
\setlength{\tabcolsep}{0.5em} 
{\renewcommand{\arraystretch}{1.3}
\begin{tabular}{cc|cc|cc}
\textbf{B-1} & \textbf{R-L} & \textbf{N-P} & \textbf{N-P Human} & \textbf{N-A} & \textbf{N-A Human} \\ \hline
82.3 & 81.08 & 94.0 & 90.0 & 65.0 & 69.0
\end{tabular}
}
\caption{\label{nli_metric_results}
Metrics on randomly sampled 100 development set example from MS MARCO 2.1 with predictions from the AnswerBART model on gold passages. N-P Human and N-A Human are the metrics computed with human annotation.
}
\vspace{-4mm}
\end{table}

\section{Conclusion}

We presented AnswerBART, an end-to-end model for conversational answer generation. AnswerBART achieves state-of-the-art results on MS MARCO 2.1 NLGEN and NarrativeQA. We evaluated the factuality of generated responses using NLI entailment showing that AnswerBART exhibits hallucinations in 3-4\% of cases.

In the future, we want to explore techniques for reducing hallucinations using constrained decoding and answer over-generation combined with answer selection~\citep{baheti2020fluent}.

\appendix

\section{Supplementary Material}

\subsection{Hardware}

We use 8 NVIDIA V100 GPUs with 32GB memory. Training AnswerBART-large takes around 2 days on 8 GPUs, and AnswerBART-base takes around 1 day on 8 GPUs. For both AnswerBART large and base, answer generation takes on average 2 seconds with 10 passages on 1 NVIDIA V100 GPU.

\subsection{Dataset Links}

\begin{itemize}
\item MS MARCO 2.1 \url{https://microsoft.github.io/msmarco/}
\item NarrativeQA \url{https://github.com/deepmind/narrativeqa}
\end{itemize}

\subsection{Hyperparamters}

We use the hyperparamters in Table~\ref{trainign_hyper} for training AnswerBART.

\begin{table}[!ht]
\centering
\small
\setlength{\tabcolsep}{0.5em} 
{\renewcommand{\arraystretch}{1.2}
\begin{tabular}{ll}
\hline
\textbf{Hyperparameter} & \textbf{Value} \\ \hline
Epochs & 5 \\
Warmup steps & 10\% \\
Weight decay & Exponential $\gamma = 0.95$ \\
Optimizer & Stochastic Weight Averaging (SWA) \\
Learning rate & 6e-4 \\
SWA learning rate & 2e-4 \\
SWA frequency & 227 \\
Batch size & 16 \\
Attention dropout & 0.1 \\
Dropout & 0.1 
\end{tabular}
}
\caption{\label{trainign_hyper}
Training hyperparamters
}
\end{table}

\subsection{AnswerBART Generated Examples}

Table~\ref{answer_bart_results} on the next pages lists examples of AnswerBART generation from MS MARCO together with NLI on Passage and NLI on Answer predictions.

Note the queries ``what airlines fly to flagstaff flg'' and ``cost of parking at bna airport``, in both cases AnswerBART hallucinated geographical locations.

\begin{table*}[!ht]
\centering
\small
\setlength{\tabcolsep}{0.5em} 
{\renewcommand{\arraystretch}{1.2}
\begin{tabular}
{p{0.15\linewidth}p{0.24\linewidth}p{0.15\linewidth}p{0.15\linewidth}p{0.05\linewidth}p{0.05\linewidth}}
\hline
\textbf{Question} & \textbf{Gold Passage} & \textbf{Gold Answer} & \textbf{Gen Answer} & \textbf{N-P} & \textbf{N-A} \\ \hline

when did sir arthur conan doyle die & Sir Arthur Conan Doyle (May 22nd, 1859 to \underline{July 7th, 1930}) was 71 when he died of a heart attack & Sir Arthur Conan Doyle died on July 7, 1930 & Sir Arthur Conan Doyle died on July 7, 1930 & entail & entail \\

where neon found in nature & In nature, the chemical element neon is found in its gaseous state only. It is found in the Earth's \underline{atmosphere} in trace amounts. & In nature, neon is found in the Earth's atmosphere & In nature, neon is found in its gaseous state only & entail & neutral \\ 

what is soya foods & Soy is a great \underline{protein source}, especially for vegetarians who do not eat meat. Soy provides the body with calcium and iron as well as other important vitamins and minerals. Soy can be found in plant-based foods including soybeans and products made from soybeans like soymilk and tofu & The soya foods are a great protein source, especially for vegetarians who do not eat meat. & Soybeans and products made from soybeans like soymilk and tofu are soya foods & entail & neutral \\

what airlines fly to flagstaff flg & Flagstaff Pulliam Airport (FLG) Flagstaff Airport is the closest small commercial airport to Flagstaff, just a 5-mile drive from downtown. Flagstaff is currently serviced daily by one commercial airline: \underline{US Airway Express} flying into Flagstaff from exclusively Phoenix. & US Airway Express is an airline that flies to Flagstaff &  US Airway Express flies to Flagstaff, \textit{Florida} & contr. & contr. \\

cost of parking at bna airport & In 1990, Executive Travel \& Parking opened its doors as Nashville International Airport’s first valet airport parking service. Our daily parking rate of \$10.50 per day is considerably less than onsite airport valet parking (\$24.00). & The cost of parking at BNA airport is \$10.50 per day & The cost of parking at \textit{Baltimore-Washington International} Airport is \$10.50 per day. & contr. & contr. \\

what converts prothrombin to thrombin &  Thrombin is produced by the enzymatic cleavage of two sites on prothrombin by \underline{activated Factor} X (Xa). The activity of factor Xa is greatly enhanced by binding to activated Factor V (Va), termed the prothrombinase complex. Prothrombin is produced in the liver and is co-translationally modified in a vitamin K-dependent reaction that converts ten glutamic acids on prothrombin to gamma-carboxyglutamic acid (Gla) & Activated Factor X converts prothrombin to thrombin. & The prothrombinase complex converts prothrombin to thrombin. & neutral & contr. \\

salad from mcdonalds calories & A super-size portion of fries at McDonald's contains 486 calories, more than double the 206 calories in a regular serving. McDonald's said of its salad that without the creamy dressing and croutons, the fat falls away and the less calorific balsamic vinegar dressing contains just 2g of fat. A spokeswoman added: Free of dressing a chicken salad has only \underline{222} calories. & At McDonald's, there are 222 calories of a chicken salad which is free of dressing & There are 206 calories in a salad from McDonald's. & contr. & contr. \\

\end{tabular}
}
\caption{\label{answer_bart_results}
MS MARCO AnswerBART Generated Examples and NLI on Passage (N-P) and Answer (N-A). Hallucinations are in italic.
}
\end{table*}



\bibliography{anthology}

\begin{thebibliography}{21}
\expandafter\ifx\csname natexlab\endcsname\relax\def\natexlab#1{#1}\fi

\bibitem[{Aly et~al.(2018)Aly, Lakhotia, Zhao, Mohit, Oguz, Arora, Gupta,
  Dewan, Nelson-Lindall, and Shah}]{aly2018pytext}
Ahmed Aly, Kushal Lakhotia, Shicong Zhao, Mrinal Mohit, Barlas Oguz, Abhinav
  Arora, Sonal Gupta, Christopher Dewan, Stef Nelson-Lindall, and Rushin Shah.
  2018.
\newblock Pytext: A seamless path from nlp research to production.
\newblock \emph{arXiv preprint arXiv:1812.08729}.

\bibitem[{Baheti et~al.(2020)Baheti, Ritter, and Small}]{baheti2020fluent}
Ashutosh Baheti, Alan Ritter, and Kevin Small. 2020.
\newblock Fluent response generation for conversational question answering.
\newblock \emph{arXiv preprint arXiv:2005.10464}.

\bibitem[{Bajaj et~al.(2018)Bajaj, Campos, Craswell, Deng, Gao, Liu, Majumder,
  McNamara, Mitra, Nguyen et~al.}]{bajaj2018ms}
Payal Bajaj, Daniel Campos, Nick Craswell, Li~Deng, Jianfeng Gao, Xiaodong Liu,
  Rangan Majumder, Andrew McNamara, Bhaskar Mitra, Tri Nguyen, et~al. 2018.
\newblock Ms marco: A human generated machine reading comprehension dataset.
\newblock \emph{arXiv preprint arXiv:1611.09268}.

\bibitem[{Clark et~al.(2020)Clark, Luong, Le, and Manning}]{clark2020electra}
Kevin Clark, Minh-Thang Luong, Quoc~V Le, and Christopher~D Manning. 2020.
\newblock Electra: Pre-training text encoders as discriminators rather than
  generators.
\newblock \emph{arXiv preprint arXiv:2003.10555}.

\bibitem[{Devlin et~al.(2018)Devlin, Chang, Lee, and
  Toutanova}]{devlin2018bert}
Jacob Devlin, Ming-Wei Chang, Kenton Lee, and Kristina Toutanova. 2018.
\newblock Bert: Pre-training of deep bidirectional transformers for language
  understanding.
\newblock \emph{arXiv preprint arXiv:1810.04805}.

\bibitem[{Einolghozati et~al.(2020)Einolghozati, Gupta, Diedrick, and
  Gupta}]{einolghozati2020sound}
Arash Einolghozati, Anchit Gupta, Keith Diedrick, and Sonal Gupta. 2020.
\newblock \href {http://arxiv.org/abs/2011.01993} {Sound natural: Content
  rephrasing in dialog systems}.

\bibitem[{Izacard and Grave(2020)}]{izacard2020leveraging}
Gautier Izacard and Edouard Grave. 2020.
\newblock Leveraging passage retrieval with generative models for open domain
  question answering.
\newblock \emph{arXiv preprint arXiv:2007.01282}.

\bibitem[{Izmailov et~al.(2018)Izmailov, Podoprikhin, Garipov, Vetrov, and
  Wilson}]{izmailov2018averaging}
Pavel Izmailov, Dmitrii Podoprikhin, Timur Garipov, Dmitry Vetrov, and
  Andrew~Gordon Wilson. 2018.
\newblock Averaging weights leads to wider optima and better generalization.
\newblock \emph{arXiv preprint arXiv:1803.05407}.

\bibitem[{Karpukhin et~al.(2020)Karpukhin, O{\u{g}}uz, Min, Wu, Edunov, Chen,
  and Yih}]{karpukhin2020dense}
Vladimir Karpukhin, Barlas O{\u{g}}uz, Sewon Min, Ledell Wu, Sergey Edunov,
  Danqi Chen, and Wen-tau Yih. 2020.
\newblock Dense passage retrieval for open-domain question answering.
\newblock \emph{arXiv preprint arXiv:2004.04906}.

\bibitem[{Ko{\v{c}}isk{\`y} et~al.(2018)Ko{\v{c}}isk{\`y}, Schwarz, Blunsom,
  Dyer, Hermann, Melis, and Grefenstette}]{kovcisky2018narrativeqa}
Tom{\'a}{\v{s}} Ko{\v{c}}isk{\`y}, Jonathan Schwarz, Phil Blunsom, Chris Dyer,
  Karl~Moritz Hermann, G{\'a}bor Melis, and Edward Grefenstette. 2018.
\newblock The narrativeqa reading comprehension challenge.
\newblock \emph{Transactions of the Association for Computational Linguistics},
  6:317--328.

\bibitem[{Lewis et~al.(2019)Lewis, Liu, Goyal, Ghazvininejad, Mohamed, Levy,
  Stoyanov, and Zettlemoyer}]{lewis2019bart}
Mike Lewis, Yinhan Liu, Naman Goyal, Marjan Ghazvininejad, Abdelrahman Mohamed,
  Omer Levy, Ves Stoyanov, and Luke Zettlemoyer. 2019.
\newblock Bart: Denoising sequence-to-sequence pre-training for natural
  language generation, translation, and comprehension.
\newblock \emph{arXiv preprint arXiv:1910.13461}.

\bibitem[{Lewis et~al.(2020)Lewis, Perez, Piktus, Petroni, Karpukhin, Goyal,
  K{\"u}ttler, Lewis, Yih, Rockt{\"a}schel et~al.}]{lewis2020retrieval}
Patrick Lewis, Ethan Perez, Aleksandara Piktus, Fabio Petroni, Vladimir
  Karpukhin, Naman Goyal, Heinrich K{\"u}ttler, Mike Lewis, Wen-tau Yih, Tim
  Rockt{\"a}schel, et~al. 2020.
\newblock Retrieval-augmented generation for knowledge-intensive nlp tasks.
\newblock \emph{arXiv preprint arXiv:2005.11401}.

\bibitem[{Liu et~al.(2019)Liu, Ott, Goyal, Du, Joshi, Chen, Levy, Lewis,
  Zettlemoyer, and Stoyanov}]{liu2019roberta}
Yinhan Liu, Myle Ott, Naman Goyal, Jingfei Du, Mandar Joshi, Danqi Chen, Omer
  Levy, Mike Lewis, Luke Zettlemoyer, and Veselin Stoyanov. 2019.
\newblock Roberta: A robustly optimized bert pretraining approach.
\newblock \emph{arXiv preprint arXiv:1907.11692}.

\bibitem[{Marcus(2020)}]{marcus2020next}
Gary Marcus. 2020.
\newblock The next decade in ai: four steps towards robust artificial
  intelligence.
\newblock \emph{arXiv preprint arXiv:2002.06177}.

\bibitem[{Nishida et~al.(2019)Nishida, Saito, Nishida, Shinoda, Otsuka, Asano,
  and Tomita}]{nishida2019multi}
Kyosuke Nishida, Itsumi Saito, Kosuke Nishida, Kazutoshi Shinoda, Atsushi
  Otsuka, Hisako Asano, and Junji Tomita. 2019.
\newblock Multi-style generative reading comprehension.
\newblock \emph{arXiv preprint arXiv:1901.02262}.

\bibitem[{Nogueira and Cho(2019)}]{nogueira2019passage}
Rodrigo Nogueira and Kyunghyun Cho. 2019.
\newblock Passage re-ranking with bert.
\newblock \emph{arXiv preprint arXiv:1901.04085}.

\bibitem[{Raffel et~al.(2019)Raffel, Shazeer, Roberts, Lee, Narang, Matena,
  Zhou, Li, and Liu}]{raffel2019exploring}
Colin Raffel, Noam Shazeer, Adam Roberts, Katherine Lee, Sharan Narang, Michael
  Matena, Yanqi Zhou, Wei Li, and Peter~J Liu. 2019.
\newblock Exploring the limits of transfer learning with a unified text-to-text
  transformer.
\newblock \emph{arXiv preprint arXiv:1910.10683}.

\bibitem[{Rajpurkar et~al.(2018)Rajpurkar, Jia, and Liang}]{rajpurkar2018know}
Pranav Rajpurkar, Robin Jia, and Percy Liang. 2018.
\newblock Know what you don't know: Unanswerable questions for squad.
\newblock \emph{arXiv preprint arXiv:1806.03822}.

\bibitem[{Tan et~al.(2017)Tan, Wei, Yang, Du, Lv, and Zhou}]{tan2017s}
Chuanqi Tan, Furu Wei, Nan Yang, Bowen Du, Weifeng Lv, and Ming Zhou. 2017.
\newblock S-net: From answer extraction to answer generation for machine
  reading comprehension.
\newblock \emph{arXiv preprint arXiv:1706.04815}.

\bibitem[{Vaswani et~al.(2017)Vaswani, Shazeer, Parmar, Uszkoreit, Jones,
  Gomez, Kaiser, and Polosukhin}]{vaswani2017attention}
Ashish Vaswani, Noam Shazeer, Niki Parmar, Jakob Uszkoreit, Llion Jones,
  Aidan~N Gomez, {\L}ukasz Kaiser, and Illia Polosukhin. 2017.
\newblock Attention is all you need.
\newblock In \emph{Advances in neural information processing systems}, pages
  5998--6008.

\bibitem[{Williams et~al.(2018)Williams, Nangia, and Bowman}]{N18-1101}
Adina Williams, Nikita Nangia, and Samuel Bowman. 2018.
\newblock \href {http://aclweb.org/anthology/N18-1101} {A broad-coverage
  challenge corpus for sentence understanding through inference}.
\newblock In \emph{Proceedings of the 2018 Conference of the North American
  Chapter of the Association for Computational Linguistics: Human Language
  Technologies, Volume 1 (Long Papers)}, pages 1112--1122. Association for
  Computational Linguistics.

\end{thebibliography}
\bibliographystyle{acl_natbib}

\end{document}